\documentclass[conference]{IEEEtran}
\IEEEoverridecommandlockouts

\usepackage{cite}
\usepackage{amsmath,amssymb,amsfonts}
\usepackage{algorithmic}
\usepackage{graphicx}
\usepackage{textcomp}

\usepackage[inline]{enumitem}
\usepackage{xcolor}

\usepackage{witharrows}

\usepackage[flushleft]{threeparttable}
\usepackage{tablefootnote}
\usepackage{array}

\newcolumntype{P}[1]{>{\centering\arraybackslash}m{#1}}
\usepackage{multicol}
\usepackage{multirow}
\usepackage{tabulary}
\usepackage{colortbl}
\definecolor{mygray}{gray}{0.90}
\usepackage{makecell}
\usepackage{booktabs}
\usepackage{subcaption}
\usepackage{stmaryrd}
\usepackage{float}
\usepackage{pifont}

\usepackage{arydshln}
\colorlet{mygray}{gray!15!white}

\usepackage[switch,columnwise]{lineno}

\usepackage{cite}

\usepackage{url,hyperref,microtype}
\hypersetup{
    colorlinks=true,
    citecolor=blue,
    linkcolor=blue,
    filecolor=magenta,      
    urlcolor=black,
}

\def\BibTeX{{\rm B\kern-.05em{\sc i\kern-.025em b}\kern-.08em
    T\kern-.1667em\lower.7ex\hbox{E}\kern-.125emX}}

\begin{document}

\title{Towards a Unified Modality-Agnostic Multimodal Framework for Cognitive Workload Assessment\\
}

\author{

\IEEEauthorblockN{Stefanos Gkikas}
\IEEEauthorblockA{\textit{Honda Research Institute Japan} \\
Wako City, Japan \\
stefanos.gkikas@jp.honda-ri.com}

\and

\IEEEauthorblockN{Christian Arzate Cruz}
\IEEEauthorblockA{\textit{Honda Research Institute Japan} \\
Wako City, Japan \\
christian.arzate@jp.honda-ri.com}

\and

\IEEEauthorblockN{Calvin Joseph}
\IEEEauthorblockA{\textit{BioSIS (Biosensing \& Intelligent Systems) Lab} \\
\textit{Centre for Intelligent Computing and Systems} \\
\textit{University of Canberra} \\
Canberra, Australia \\
calvin.joseph@canberra.edu.au}

\and

\rule{0pt}{20pt}\IEEEauthorblockN{Giorgos Giannakakis}
\IEEEauthorblockA{\textit{Department of Electronic Engineering} \\
\textit{Hellenic Mediterranean University}\\
Chania, Greece \\
ggian@hmu.gr}
\and

\and
\IEEEauthorblockN{Raul Fernandez Rojas}
\IEEEauthorblockA{\textit{BioSIS (Biosensing \& Intelligent Systems) Lab} \\
\textit{Centre for Intelligent Computing and Systems} \\
\textit{University of Canberra} \\
Canberra, Australia \\
raul.fernandezrojas@canberra.edu.au}

}

\maketitle

\begin{abstract}

Cognitive workload reflects the mental effort required during task performance and is central to the design of adaptive human-machine systems. The use of biosignals to measure cognitive workload has been extensively researched and documented; however, studies examining the effects of combining heterogeneous biosignal modalities for this purpose remain limited. To provide insight into this area, we developed a unified, modality-agnostic, hierarchical Transformer-based architecture to process heterogeneous biosignal modalities within a single model.
We use this framework in a pilot study evaluating all $31$ possible combinations of five modalities: Electrocardiogram (ECG), Electrodermal Activity (EDA), Respiration (RESP), Peripheral Oxygen Saturation (SpO\textsubscript{2}), and Electroencephalogram (EEG), under leave one subject out validation across three cognitively distinct tasks: abstract reasoning (IQ), arithmetic problem solving (MATH), and a game task (GAME).
In this pilot setting, the results suggest that: (i) EEG is the strongest single modality, ranking highest in IQ, GAME, and the pooled ALL setting, where samples from all three tasks are combined; (ii) adding more modalities does not consistently improve performance; (iii) the full five modality combination achieves the highest \textit{Average} score of $73.02\%$ on IQ and $68.08\%$ when the \textit{Average} scores are averaged over the four evaluation settings: IQ, MATH, GAME, and ALL; and (iv) the proposed method reduces model size by approximately $50\%$ compared with late fusion alternatives while maintaining lower inference time.

\end{abstract}

\begin{IEEEkeywords}
cognitive fatigue, mental workload, physiological signals, brain activity, multimodal fusion, transformer
\end{IEEEkeywords}

\section{Introduction}

Cognitive Workload is defined as the mental effort required for working memory to complete tasks. In many high-risk fields, including aviation, automotive, robotics, and automation, estimating cognitive workload in real time has significant implications for both safety and performance. The importance of detecting when an individual may require assistance due to increased cognitive workload extends beyond professional settings to health care and education. For example, early detection of increased levels of cognitive workload may provide timely support \cite{hassard_teoh_2018} \cite{anders_moontaha_2024}.
Workload assessment methods are broadly classified as subjective or objective \cite{rojas_debie_2020}. Subjective methods, such as the NASA Task Load Index (NASA-TLX), are well validated but can only be administered after task completion, thereby ruling out real-time monitoring \cite{galy_paxion_2018}.
Objective methods exploit the coupling between mental effort and autonomic and cortical nervous system activity, enabling continuous estimation from physiological sensors \cite{charles_nixon_2019}. 
There are several common physiological sensor modalities used for cognitive workload estimation, including EEG for cortical dynamics, ECG for cardiac activity, EDA and respiratory signals for autonomic responses, and blood oxygen saturation for peripheral oxygenation.
Each carries distinct and partially complementary information, and no single signal fully characterizes the multi-dimensional nature of cognitive demand \cite{hirachan_niraj_2022, healey_picard_2005}.

Despite the complementary nature of physiological signals, much of the cognitive-workload literature focuses either on single-modality models or on a limited number of predefined multimodal combinations, rather than systematically comparing the full space of modality subsets \cite{hogervorst_brouwer_2014, debie_rojas_2021, charles_nixon_2019, tao_tan_2019}. 
The lack of systematic evaluation of all modality subsets makes it difficult to provide valid guidance on modality choice when sensor resources are limited, given task variability across application domains \cite{charles_nixon_2019, tao_tan_2019, chakladar_roy_2024}. In addition, most multimodal architectures employ modality-specific feature extraction techniques and/or explicit fusion methods, such as intra-modality encoders, dual-branch networks, or decision-level fusion modules, which generally require the sensor set to be defined a priori. Therefore, they are usually designed to process and be optimized for specific input data \cite{hu_sukthankar_2024, rabbani_islam_2024, li_zhu_2025}.
This pilot study addresses these gaps by using a shared hierarchical transformer that processes any combination of five modalities without modality-specific components, and systematically evaluates all possible modality combinations using a leave-one-subject-out protocol across three cognitively distinct tasks.

\section{Related Work}
\label{related_work}

The EEG modality has been studied extensively for cognitive workload estimation, with CNN and LSTM architectures among the most common for extracting spatial and temporal representations from raw or preprocessed signals \cite{qin_bulbul_2023, pusica_kartali_2024}. However, more complex network designs have shown improved performance \cite{siddhad_roy_2024}, and systematic reviews suggest that, in general, deep learning approaches outperform feature engineering pipelines \cite{khingphai_moshfeghi_2025}. 
Transformer-based EEG and brain-activity modeling has also been explored in recent work on affective computing \cite{gkikas_cruz_eeite_cwl_2026, gkikas_arzate_eeite_pain_2026, gkikas_arzate_pain_icmi_2026}.
Furthermore, Transformer and Conformer architectures have also been applied to EEG-based workload estimation, achieving competitive results at low computational cost \cite{kostas_ruber_2020, song_zheng_2023}.
Physiological signals (ECG, EDA, RESP, and SpO\textsubscript{2}) provide complementary information that reflects the autonomic nervous system's responses to cognitive demand. Heart rate variability and electrodermal activity have been shown to discriminate between workload levels during real-world tasks \cite{healey_picard_2005}, and multimodal wearable datasets have confirmed that ECG and EDA are reliable indicators of stress and cognitive load \cite{schmidt_ring_2018, rojas_debie_2020}. The relative contribution of individual peripheral modalities across different cognitive paradigms, however, remains inadequately evaluated. Recent studies have also examined EDA, respiration, ECG, and general biosignal embedding models for physiological-state assessment~\cite{gkikas_kyprakis_eda_2025,gkikas_kyprakis_resp_2025,farmani_bargshady_2025,gkikas_tiny_2025}.

Multimodal physiological fusion has been most thoroughly studied in EEG--fNIRS configurations, where combining cortical electrophysiological activity and hemodynamic responses often improves performance over either modality independently \cite{debie_rojas_2021, saadati_nelson_2020, mathews_hirachan_2024, bunterngchit_wang_2024}. 
Attention-based fusion, mutual information feature selection, and joint feature learning have also been used to improve cross-modal alignment and cognitive state decoding \cite{deligani_borgheai_2021, li_zhu_2025}. Related work has further examined heterogeneous physiological fusion, foundation-model designs, compact biosignal representations, facial spatiotemporal representation learning, and co-speech gesture prediction \cite{bargshady_aziz_2025,khan_chetty_2026,gkikas_rojas_painformer_2025,gkikas_tiny_2025,gkikas_reface_2026,vazquez_cruz_gkikas_2026}. In most cases, these methods use modality-specific branches and evaluate a single fixed combination, which limits their applicability when sensor availability varies.


\section{Methodology}
\label{sec:methodology}

\subsection{Data Collection}
\label{ssec:data_collection}

\begin{figure*}
\begin{center}
\includegraphics[scale=0.141]{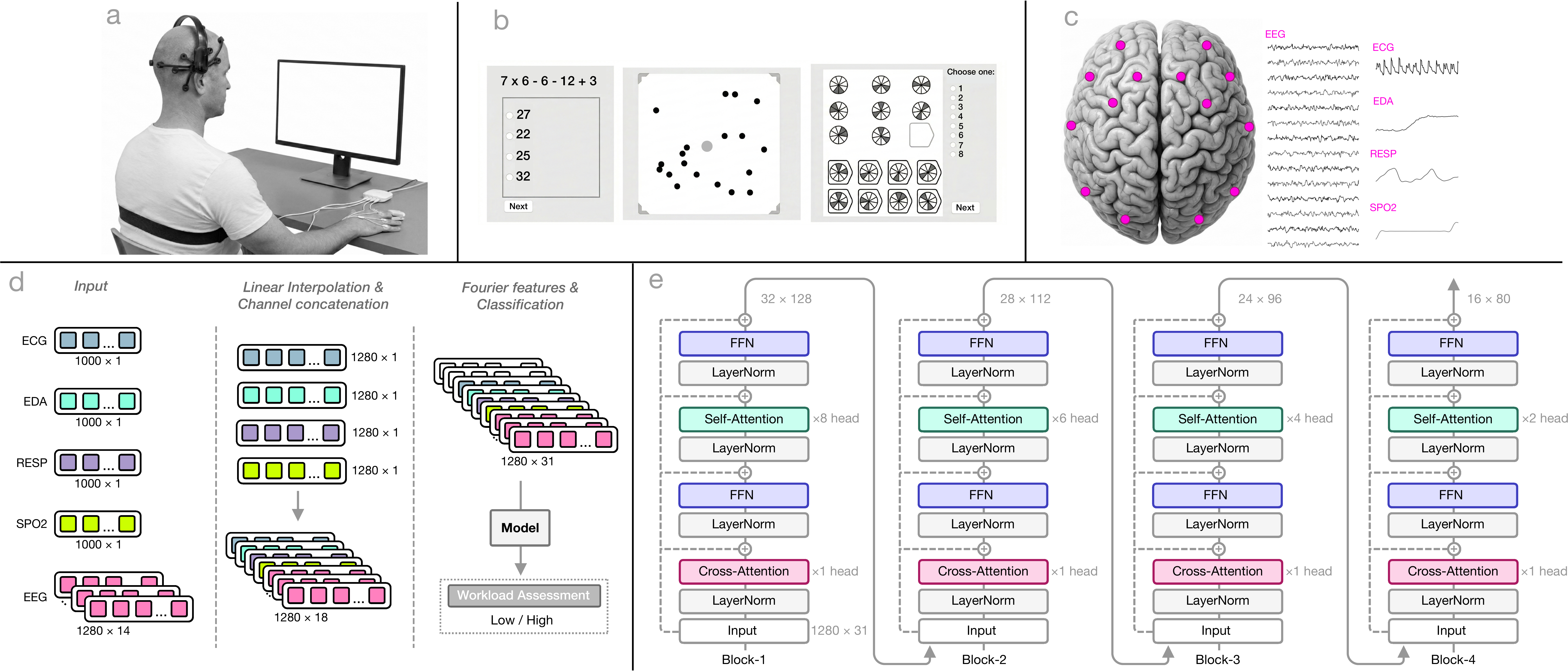}
\end{center}
\caption{Overview of the proposed multimodal cognitive workload assessment framework.
\textbf{(a)}~Experimental setup for physiological data acquisition.
\textbf{(b)}~Workload-inducing task paradigms: abstract reasoning (IQ),
arithmetic problem solving (MATH), and game-based interaction (GAME),
each presented under low- and high-workload conditions.
\textbf{(c)}~The five recorded physiological modalities (EEG: $14$ channels;
ECG, EDA, RESP, SpO\textsubscript{2}: 1 channel each) with their raw signal traces.
\textbf{(d)}~Multimodal input construction: peripheral biosignals are linearly
interpolated to the EEG temporal length (1280 samples) and concatenated along the
channel axis, producing a fused tensor of $1280\!\times\!18$, which expands to
$1280\!\times\!31$ after Fourier positional features are appended.
\textbf{(e)}~The hierarchical four-block architecture. The fused token sequence
is processed through cross-attention and repeated latent self-attention, while
latent vectors and dimensionality decrease progressively across blocks:
Block-1 ($32\!\times\!128$), Block-2 ($28\!\times\!112$),
Block-3 ($24\!\times\!96$), Block-4 ($16\!\times\!80$), producing the final
binary workload prediction (\textit{Low\,/\,High}).}
\label{overview}
\end{figure*}

Eleven participants ($6$ male and $5$ female; ages from $20$ to $40$ years; mean age $25 \pm 5.5$) participated in the study. None reported neurological disorders or use of substances affecting the nervous system on the day of the experiment. All participants provided written informed consent, and the study protocol was approved by the Institutional Review Board. 
EEG was recorded with an \textit{EMOTIV EPOC} wireless headset ($14$ channels: AF3, F7, F3, FC5, T7, P7, O1, O2, P8, T8, FC6, F4, F8, AF4) at $128$ Hz and preprocessed in EEGLAB, where ICA was used to identify and remove components dominated by ocular or muscular artifacts.
Peripheral physiological signals (ECG, EDA, RESP, and SpO\textsubscript{2}) were simultaneously acquired with \textit{PLUX Biosignals} sensors at $100$\, Hz, as illustrated in Fig.~\ref{overview}(a). 
Participants performed $3$ different cognitive tasks. Each of these tasks was presented in $2$ difficulty levels. They included a mathematical problem-solving test (MATH), Raven's Progressive Matrices (for abstract reasoning) (IQ), and an open-source video game (GAME) where participants controlled a ball that bounced around a maze.
Task order followed a randomized orthogonal design; each $2$-minute activity was followed by a NASA-TLX subjective workload assessment, as depicted in Fig.~\ref{overview}(b), and a $1$-minute rest period. Binary class labels correspond to the predefined difficulty level of each task (easy/hard); NASA-TLX scores were collected as a subjective workload verification and are not used as classification labels. The dataset is balanced across both workload classes and all modalities, with each participant providing $12$ samples per difficulty level across three tasks, yielding $792$ samples per modality and $3{,}960$ samples in total.

\subsection{Multimodal Fusion and Tokenization}
\label{ssec:fusion}

Five physiological modalities are fused into a single token sequence before being processed by the transformer: ECG, EDA, RESP, SpO\textsubscript{2}, and EEG, as shown in Fig.~\ref{overview}(c).
The 10-second windows yield $L_{\text{bio}} = 1000$ data points for peripheral signals and $L_{\text{EEG}} = 1280$ data points across $C_{\text{EEG}} = 14$ channels for EEG.

\subsubsection{Temporal alignment}
When EEG is included in the selected modality set $\mathcal{M}$, each
peripheral signal
$\mathbf{x}^{(m)} \in \mathbb{R}^{L_{\text{bio}}}$,
$m \in \mathcal{M}_{\text{bio}}$,
is resampled to $L_{\text{EEG}}$ via linear interpolation:
\begin{equation}
  \hat{x}^{(m)}_{j}
    = x^{(m)}_{i}
    + \!\left(\frac{j\,(L_{\text{bio}}-1)}{L_{\text{EEG}}-1} - i\right)
      \!\left(x^{(m)}_{i+1} - x^{(m)}_{i}\right),
  \label{eq:interp}
\end{equation}
where
$i = \bigl\lfloor j(L_{\text{bio}}{-}1)/(L_{\text{EEG}}{-}1)
\bigr\rfloor$
and $j = 0,\ldots,L_{\text{EEG}}{-}1$.
When the EEG is absent, all signals retain their native length $L_{\text{bio}}$ and no interpolation is applied. The interpolation step is used only to align temporal sequence lengths before channel stacking. No frequency-domain biosignal features are extracted, and the Fourier features below encode temporal position rather than signal spectra.

\subsubsection{Channel-stack fusion}
The multimodal input tensor is formed by concatenating all selected modalities along the channel dimension:
\begin{equation}
  \mathbf{X}
    = \bigl[\hat{\mathbf{x}}^{(1)}
            \;\|\; \cdots
            \;\|\; \hat{\mathbf{x}}^{(|\mathcal{M}|)}\bigr]
    \in \mathbb{R}^{C \times L},
  \label{eq:fusion}
\end{equation}
where
$C = C_{\text{EEG}}\,\mathbf{1}[\text{EEG}\in\mathcal{M}]
     + |\mathcal{M}_{\text{bio}}|$
and $L = L_{\text{EEG}}$ if EEG is present, otherwise $L_{\text{bio}}$.
No modality-specific branches or encoders are used; all channels are treated uniformly in a shared representation space.
For the full pentamodal configuration, $C{=}18$ and $L{=}1280$,
as illustrated in Fig.~\ref{overview}(d).
The unimodal configuration corresponds to a special degenerate case of \eqref{eq:fusion} with
$|\mathcal{M}|{=}1$.

\subsubsection{Input normalization}
Prior to tokenization, each channel of $\mathbf{X}$ is normalized over the temporal dimension via a learnable Layer Normalization:
\begin{equation}
  \mathrm{LN}(\mathbf{x})
    = \frac{\mathbf{x} - \mu_{\mathbf{x}}}
           {\sqrt{\sigma^{2}_{\mathbf{x}} + \varepsilon}}
      \odot \boldsymbol{\gamma}
      + \boldsymbol{\beta},
  \label{eq:ln_input}
\end{equation}
where $\mu_{\mathbf{x}}$ and $\sigma^{2}_{\mathbf{x}}$ are the mean and
variance computed over the $L$ temporal positions of that channel, and
$\boldsymbol{\gamma}, \boldsymbol{\beta} \in \mathbb{R}^{L}$ are learnable
affine parameters. This operation standardizes signal amplitudes across
physiologically heterogeneous modalities before they enter the transformer.

\subsubsection{Tokenization}
Each temporal position $n = 1,\ldots, N$ (where $N{=}L$) is treated as a token whose feature vector combines the raw signal values with
Fourier-based positional encodings~\cite{jaegle_gimeno_2021}.
Each position is mapped to a normalized coordinate
$p_{n} = 2(n{-}1)/(N{-}1) - 1 \in [-1,1]$.
The Fourier encoding with $K$ frequency bands and maximum frequency
$f_{\max}$ is defined as:
\begin{multline}
  \boldsymbol{\gamma}(p)
    = \bigl[\sin(\pi f_{1} p),\;\cos(\pi f_{1} p),\;\ldots,\;\\
             \sin(\pi f_{K} p),\;\cos(\pi f_{K} p),\;p\bigr],
  \label{eq:fourier}
\end{multline}
where the frequencies are spaced geometrically on $[1,\,f_{\max}/2]$:
\begin{equation}
  f_{k} = 2^{\,(k-1)\,\log_{2}(f_{\max}/2)\,/\,(K-1)},
  \qquad k = 1,\ldots,K.
  \label{eq:freqs}
\end{equation}
The complete token matrix is formed by concatenating data channels and
positional features at every position:
\begin{equation}
  \mathbf{T}
    = \bigl[\mathbf{X}^{\top} \;\|\; \boldsymbol{\Gamma}\bigr]
    \in \mathbb{R}^{N \times C'},
  \label{eq:tokens}
\end{equation}
where $\boldsymbol{\Gamma}\in\mathbb{R}^{N\times(2K+1)}$ stacks the
positional encodings row-wise and $C' = C + (2K+1)$.
With $K{=}6$ and $f_{\max}{=}10$, the positional dimension is
$2K{+}1{=}13$, so $C'{=}C{+}13$.
For the pentamodal case ($C{=}18$) this gives
$\mathbf{T}\in\mathbb{R}^{1280\times31}$,
consistent with Fig. \ref{overview}(d).

\subsection{Hierarchical Transformer Architecture}
\label{ssec:arch}

The model processes $\mathbf{T}\in\mathbb{R}^{N\times C'}$ through four sequential transformer blocks, illustrated in Fig.~\ref{overview}(e).
Both the number of latent vectors $M_{\ell}$ and the latent dimensionality $d_{\ell}$ decrease across blocks, progressively compressing the representation towards a compact encoding for classification.
The per-block hyperparameters are provided in Table~\ref{table:architecture} (Appendix). The resulting model is lightweight, comprising $5.61$M parameters and requiring $0.36$~GFLOPs per forward pass.

\subsubsection{Latent initialization}
A global latent array $\mathbf{E}\in\mathbb{R}^{M_{1}\times d_{1}}$ of
learnable vectors is shared across the batch. At the entry to Block-1, the latent state is initialized as:
\begin{equation}
\mathbf{Z}^{(1)} = \mathbf{E} \in \mathbb{R}^{M_{1}\times d_{1}}.
\label{eq:init}
\end{equation}

\subsubsection{Cross-attention}
Each block begins with a cross-attention operation in which the latent array
queries the full token sequence $\mathbf{T}$, using pre-layer normalization
and an additive residual connection:
\begin{align}
\mathbf{Z} &\leftarrow \mathbf{Z}
+ \mathrm{MHA}_{\text{cr}}\!\bigl(\mathrm{LN}(\mathbf{Z}),\;
                           \mathbf{T}\bigr),
\label{eq:cross_attn} \\[2pt]
\mathbf{Z} &\leftarrow \mathbf{Z}
+ \mathrm{FFN}\!\bigl(\mathrm{LN}(\mathbf{Z})\bigr),
\label{eq:cross_ff}
\end{align}
where $\mathrm{MHA}_{\text{cr}}$ uses the latents as queries and the token
matrix as keys and values, with $H_{\text{cr}}{=}1$ head in every block.
Multi-head attention is computed as:
\begin{equation}
\mathrm{MHA}(\mathbf{Q},\mathbf{K},\mathbf{V})
= \mathrm{Concat}(\mathrm{head}_{1},\ldots, \mathrm{head}_{H})\,\mathbf{W}^{O},
\label{eq:mha}
\end{equation}
\begin{equation}
\mathrm{head}_{h}
= \mathrm{softmax}\!\!\left(
\frac{\mathbf{Q}\mathbf{W}^{Q}_{h}
\bigl(\mathbf{K}\mathbf{W}^{K}_{h}\bigr)^{\!\top}}
{\sqrt{d_{h}}}
\right)\mathbf{V}\mathbf{W}^{V}_{h}.
\label{eq:attn_head}
\end{equation}
The single cross-attention head keeps the cost of reading from the token
sequence proportional to $\mathcal{O}(M_{\ell} N)$.
The feedforward network uses a GELU-activated two-layer projection:
\begin{equation}
\mathrm{FFN}(\mathbf{z})
= \mathbf{W}_{2}\;\sigma_{\mathrm{GELU}}(\mathbf{W}_{1}\mathbf{z}).
\label{eq:ffn}
\end{equation}

\subsubsection{Latent self-attention}
Following cross-attention, the latent array undergoes $R_{\ell}$ rounds of
self-attention within the latent space at cost $\mathcal{O}(M_{\ell}^{2})$.
For each round $r = 1,\ldots,R_{\ell}$:
\begin{align}
\mathbf{Z} &\leftarrow \mathbf{Z}
+ \mathrm{MHA}_{\text{la}}\!\bigl(\mathrm{LN}(\mathbf{Z}),\;
                               \mathrm{LN}(\mathbf{Z})\bigr),
\label{eq:self_attn} \\[2pt]
\mathbf{Z} &\leftarrow \mathbf{Z}
+ \mathrm{FFN}\!\bigl(\mathrm{LN}(\mathbf{Z})\bigr).
\label{eq:self_ff}
\end{align}
All attention and feedforward operations use pre-layer normalization with
additive residual connections, which stabilize gradient flow across the
depth of the model.

\subsubsection{Between-block transition}
After block $\ell$ the latent state
$\mathbf{Z}^{(\ell)}_{\text{out}}\in\mathbb{R}^{M_{\ell}\times d_{\ell}}$
undergoes a two-stage transition. First, a linear projection adapts the
channel dimension:
\begin{equation}
\tilde{\mathbf{Z}}^{(\ell)}
= \mathbf{Z}^{(\ell)}_{\text{out}}\,\mathbf{W}^{(\ell)}_{\text{proj}}
\in \mathbb{R}^{M_{\ell}\times d_{\ell+1}},
\label{eq:proj}
\end{equation}
where $\mathbf{W}^{(\ell)}_{\text{proj}}\in
\mathbb{R}^{d_{\ell}\times d_{\ell+1}}$. Second, the number of latent vectors is reduced via one-dimensional adaptive average pooling along the latent axis:
\begin{equation}
\mathbf{Z}^{(\ell+1)}
= \mathrm{AdaptiveAvgPool}_{M_{\ell}\to M_{\ell+1}}\!
\bigl(\tilde{\mathbf{Z}}^{(\ell)}\bigr)
\in \mathbb{R}^{M_{\ell+1}\times d_{\ell+1}}.
\label{eq:pool}
\end{equation}
This joint reduction of channel dimension and latent count enforces an
information bottleneck at every block boundary, requiring each successive
block to work from a more compact representation.

\subsubsection{Classification head}
After Block-4, the final latent state
$\mathbf{Z}^{(4)}\in\mathbb{R}^{M_{4}\times d_{4}}$ is averaged over latent vectors:
\begin{equation}
  \mathbf{z}_{\text{cls}}
    = \frac{1}{M_{4}}\sum_{m=1}^{M_{4}}\mathbf{Z}^{(4)}_{m}
    \in \mathbb{R}^{d_{4}}.
  \label{eq:mean_pool}
\end{equation}
A final Layer Normalization followed by a linear projection produces the
class logits:
\begin{equation}
  \hat{\mathbf{y}}
    = \mathbf{W}_{\text{cls}}\,\mathrm{LN}(\mathbf{z}_{\text{cls}})
    + \mathbf{b}_{\text{cls}} \in \mathbb{R}^{N_{c}},
  \label{eq:classifier}
\end{equation}
where $N_{c} = 2$ for binary workload classification.

\subsection{Augmentation, Regularization, and Training}
\label{ssec:training}
Three channel-synchronized augmentations are applied during training: \textit{polarity inversion}, \textit{additive Gaussian noise}, and \textit{temporal masking}. Additional regularization includes label smoothing and dropout. Models are trained with \textit{AdamW} and a cosine-annealing schedule. Full hyperparameters are provided in Table \ref{table:training} (Appendix).

\section{Experimental Evaluation \& Results}
This section reports results for all possible modality combinations, ranging from single-modality models to the full five-modality set (ECG, EDA, RESP, SpO$_2$, EEG), across three tasks (IQ, MATH, GAME) and a cross-task setting (ALL), under a leave-one-subject-out (LOSO) protocol. 
Each task is framed as a binary classification problem using the predefined easy/hard task-difficulty conditions as controlled workload-inducing labels, rather than direct subjective NASA-TLX targets.
Performance is reported using accuracy, precision, and F1, together with their arithmetic mean (\textit{Average}), which serves as the primary criterion. \textit{Average} provides a compact summary that favors modality combinations performing consistently across all three reported metrics.

\subsection{Unimodal Assessment}
Table~\ref{table:single_half} reports unimodal performance across all four task settings. EEG is the best single modality in three of four settings, achieving the highest averages of $69.22\%$ (IQ), $72.27\%$ (GAME), and $63.96\%$ (ALL). For GAME, its advantage over RESP (the second-best modality) is $72.27\%$ vs.\ $70.29\%$. In individual metrics, EEG achieves $71.97\%$ accuracy, $73.53\%$ precision, and $71.32\%$ F1, while RESP reaches $69.32\%$, $73.99\%$, and $67.56\%$. RESP shows slightly higher precision, but EEG's higher accuracy and F1 score result in a higher overall \textit{Average}. 
This is consistent with EEG capturing task-relevant cortical dynamics of the gaming task more reliably across both workload levels.

ECG is the weakest-performing modality across all three tasks. It yields the lowest performance in the ALL setting, achieving an \textit{Average} of $46.52 \%$, a precision of $32.51 \%$, failing to achieve chance-level discrimination. The low quality of this signal suggests that cardiac activity lacks the specificity needed to accurately discriminate task workloads. For MATH, the RESP yields the largest \textit{Average} of $65.64\%$ and slightly surpasses EEG's ($65.22\%$). Both modalities had similar accuracy ($64.77\%$), but RESP presented higher precision ($70.35\%$) than EEG ($67.19\%$), while EEG had a higher F1 ($63.69\%$) than RESP ($61.79\%$). These differences indicate that both modalities capture distinct characteristics of the MATH workload; RESP captures changes in overall physiological load with greater specificity, whereas EEG offers a better-balanced trade-off between precision and recall.
EDA and SpO$_2$  represent a middle ground in terms of performance, with SpO$_2$ being generally more stable. All modalities showed that the worst scores were obtained in the ALL condition, since combining physiological data from three separate cognitive tasks increases inter-task variability within the same class. 
This pattern is consistent across all modalities.

\begin{table}
\scriptsize
\caption{Unimodal performance (\textit{low vs high}) across all task settings. \textit{Average}: arithmetic mean of Accuracy, Precision, and F1, used as the primary performance criterion throughout. \textbf{Bold} marks the highest \textit{Average} within each task; \underline{underline} marks the second-highest.}
\label{table:single_half}
\begin{center}
\begin{threeparttable}
\begin{tabular}{P{0.5cm} P{0.9cm} P{0.9cm} P{0.9cm} P{0.4cm} P{1.0cm}}
\toprule
\multirow[c]{3}{*}{Task} &
\multirow[c]{3}{*}{Modality} &
\multicolumn{4}{c}{Performance} \\
\cmidrule(lr){3-6}
&  & Accuracy & Precision & F1 &\textit{Average} \\
\midrule
\midrule
IQ &ECG                  &55.68 &59.53 &48.12 &\textit{54.44}\\\hdashline
IQ &EDA                  &63.26 &68.92 &60.52 &\textit{64.23}\\\hdashline
IQ &RESP                 &64.39 &71.75 &60.77 &\textit{\underline{65.64}}\\\hdashline
IQ &SpO\textsubscript{2} &64.02 &67.65 &59.73 &\textit{63.80}\\\hdashline
IQ &EEG                  &68.18 &74.12 &65.37 &\textit{\textbf{69.22}}\\\midrule

MATH &ECG                  &60.23 &63.09 &53.54 &\textit{58.95}\\\hdashline
MATH &EDA                  &59.47 &64.72 &56.18 &\textit{60.12}\\\hdashline
MATH &RESP                 &64.77 &70.35 &61.79 &\textit{\textbf{65.64}}\\\hdashline
MATH &SpO\textsubscript{2} &62.88 &67.54 &60.91 &\textit{63.78}\\\hdashline
MATH &EEG                  &64.77 &67.19 &63.69 &\textit{\underline{65.22}}\\\midrule

GAME &ECG                  &60.23 &67.00 &54.18 &\textit{60.47}\\\hdashline
GAME &EDA                  &64.02 &65.29 &60.32 &\textit{63.21}\\\hdashline
GAME &RESP                 &69.32 &73.99 &67.56 &\textit{\underline{70.29}}\\\hdashline
GAME &SpO\textsubscript{2} &62.50 &65.49 &58.01 &\textit{61.67}\\\hdashline
GAME &EEG                  &71.97 &73.53 &71.32 &\textit{\textbf{72.27}}\\\midrule

ALL &ECG                  &56.44 &32.51 &50.60 &\textit{46.52}\\\hdashline
ALL &EDA                  &56.82 &60.98 &49.99 &\textit{55.93}\\\hdashline
ALL &RESP                 &59.60 &60.14 &55.94 &\textit{\underline{58.56}}\\\hdashline
ALL &SpO\textsubscript{2} &55.56 &57.25 &52.70 &\textit{55.17}\\\hdashline
ALL &EEG                  &63.76 &65.52 &62.61 &\textit{\textbf{63.96}}\\
\bottomrule
\end{tabular}
\begin{tablenotes}[para,flushleft]
\scriptsize
\end{tablenotes}
\end{threeparttable}
\end{center}
\end{table}

\subsection{Bimodal Combinations}

Table~\ref{table:bimodal} reports performance for all $10$ pairwise modality combinations. EEG-containing pairs achieve the highest \textit{Average} in most settings. For IQ, RESP$+$EEG obtains the best bimodal \textit{Average} of $72.00\%$ ($71.59\%$ accuracy, $73.49\%$ precision, $70.93\%$ F1), a gain of $+2.78$ points over the best unimodal result. The improvement is consistent across all three metrics, indicating complementarity between brain and respiratory signals for abstract reasoning. For ALL, SpO$_2+$EEG achieves an average of $64.00\%$, a marginal $+0.04$ improvement over unimodal EEG, with a slight F1 gain ($62.97\%$ vs.\ $62.61\%$), while accuracy and precision remain nearly flat. For MATH, ECG$+$EEG achieves an average of $65.85\%$, only $+0.21$ above the unimodal RESP baseline, indicating limited multimodal benefit for arithmetic workload.
For GAME and ALL settings, the EEG-containing pairs perform worse than the unimodal EEG approach. The best pair, SpO$_2+$EEG, achieves an average of $68.59\%$, a decrease of $-3.68$ points relative to unimodal EEG alone ($72.27\%$). SpO$_2+$EEG reaches $68.94\%$ accuracy and $71.50\%$ precision, still close to unimodal EEG, but F1 drops to $65.33\%$ vs.\ $71.32\%$. This suggests that adding peripheral channels to the shared token sequence disrupts the EEG-based representation, rather than complementing it. 
The GAME task likely relies more strongly on EEG-derived information, making unimodal EEG an effective input modality.
Notably, among pairs that do not include EEG, ECG$+$RESP achieves $65.84\%$ for MATH, nearly matching the best EEG-containing pair, driven by strong precision ($72.45\%$). This indicates that cardiovascular and respiratory signals in isolation carry meaningful information about arithmetic workload.

\begin{table}
\scriptsize
\caption{Bimodal performance across all task settings.}
\label{table:bimodal}
\begin{center}
\begin{threeparttable}
\begin{tabular}{P{0.5cm} P{1.6cm} P{0.9cm} P{0.9cm} P{0.4cm} P{1.0cm}}
\toprule
\multirow[c]{3}{*}{Task} &
\multirow[c]{3}{*}{Modality} &
\multicolumn{4}{c}{Performance} \\
\cmidrule(lr){3-6}
&  & Accuracy & Precision & F1 &\textit{Average} \\
\midrule
\midrule
IQ &ECG, EDA                  &63.64 &64.23 &60.84 &\textit{62.90}\\\hdashline
IQ &ECG, RESP                 &62.88 &66.84 &60.60 &\textit{63.44}\\\hdashline
IQ &ECG, SpO\textsubscript{2} &65.15 &72.38 &61.53 &\textit{66.35}\\\hdashline
IQ &ECG, EEG                  &67.42 &71.98 &65.37 &\textit{68.26}\\\hdashline
IQ &EDA, RESP                 &64.02 &69.09 &60.88 &\textit{64.66}\\\hdashline
IQ &EDA, SpO\textsubscript{2} &65.53 &69.43 &63.58 &\textit{66.18}\\\hdashline
IQ &EDA, EEG                  &70.45 &72.10 &69.79 &\textit{\underline{70.78}}\\\hdashline
IQ &RESP, SpO\textsubscript{2}&65.15 &70.98 &62.11 &\textit{66.08}\\\hdashline
IQ &RESP, EEG                 &71.59 &73.49 &70.93 &\textit{\textbf{72.00}}\\\hdashline
IQ &SpO\textsubscript{2}, EEG &69.70 &72.31 &68.66 &\textit{70.22}\\\midrule

MATH &ECG, EDA                  &63.64 &65.89 &62.43 &\textit{63.99}\\\hdashline
MATH &ECG, RESP                 &64.39 &72.45 &60.68 &\textit{\underline{65.84}}\\\hdashline
MATH &ECG, SpO\textsubscript{2} &62.12 &63.81 &61.11 &\textit{62.35}\\\hdashline
MATH &ECG, EEG                  &64.77 &70.70 &62.08 &\textit{\textbf{65.85}}\\\hdashline
MATH &EDA, RESP                 &64.02 &70.97 &61.09 &\textit{65.36}\\\hdashline
MATH &EDA, SpO\textsubscript{2} &60.23 &63.47 &55.64 &\textit{59.78}\\\hdashline
MATH &EDA, EEG                  &61.74 &68.24 &55.97 &\textit{61.98}\\\hdashline
MATH &RESP, SpO\textsubscript{2}&65.15 &67.07 &64.18 &\textit{65.47}\\\hdashline
MATH &RESP, EEG                 &61.36 &68.00 &57.42 &\textit{62.26}\\\hdashline
MATH &SpO\textsubscript{2}, EEG &60.23 &69.59 &54.89 &\textit{61.57}\\\midrule

GAME &ECG, EDA                  &63.26 &64.97 &59.80 &\textit{62.68}\\\hdashline
GAME &ECG, RESP                 &65.50 &71.17 &62.72 &\textit{66.46}\\\hdashline
GAME &ECG, SpO\textsubscript{2} &62.12 &70.67 &57.64 &\textit{63.48}\\\hdashline
GAME &ECG, EEG                  &68.94 &68.05 &66.82 &\textit{67.94}\\\hdashline
GAME &EDA, RESP                 &64.77 &69.41 &62.58 &\textit{65.59}\\\hdashline
GAME &EDA, SpO\textsubscript{2} &61.74 &67.16 &59.03 &\textit{62.64}\\\hdashline
GAME &EDA, EEG                  &66.67 &65.58 &62.04 &\textit{64.76}\\\hdashline
GAME &RESP, SpO\textsubscript{2}&64.02 &65.06 &63.11 &\textit{64.06}\\\hdashline
GAME &RESP, EEG                 &67.80 &68.35 &64.78 &\textit{\underline{66.98}}\\\hdashline
GAME &SpO\textsubscript{2}, EEG &68.94 &71.50 &65.33 &\textit{\textbf{68.59}}\\\midrule

ALL &ECG, EDA                  &55.68 &62.55 &49.19 &\textit{55.81}\\\hdashline
ALL &ECG, RESP                 &56.57 &58.80 &53.90 &\textit{56.42}\\\hdashline
ALL &ECG, SpO\textsubscript{2} &55.43 &54.47 &49.00 &\textit{52.97}\\\hdashline
ALL &ECG, EEG                  &64.02 &62.73 &61.95 &\textit{62.90}\\\hdashline
ALL &EDA, RESP                 &56.57 &61.60 &52.69 &\textit{56.95}\\\hdashline
ALL &EDA, SpO\textsubscript{2} &56.06 &59.63 &51.53 &\textit{55.74}\\\hdashline
ALL &EDA, EEG                  &63.13 &61.57 &60.90 &\textit{61.87}\\\hdashline
ALL &RESP, SpO\textsubscript{2}&56.82 &62.32 &52.47 &\textit{57.20}\\\hdashline
ALL &RESP, EEG                 &63.01 &64.89 &61.98 &\textit{\underline{63.29}}\\\hdashline
ALL &SpO\textsubscript{2}, EEG &63.76 &65.28 &62.97 &\textit{\textbf{64.00}}\\
\bottomrule
\end{tabular}
\begin{tablenotes}[para,flushleft]
\scriptsize
\end{tablenotes}
\end{threeparttable}
\end{center}
\end{table}

\subsection{Trimodal Combinations}

Table~\ref{table:trimodal} presents the results for the $3$-modality combinations. For IQ, the best trimodal configuration, ECG$+$RESP$+$EEG, achieves $69.10\%$ on average, a drop of $-2.90$ points from the bimodal peak of $72.00\%$. Precision declines from $73.49\%$ to $72.35\%$, and F1 drops more sharply from $70.93\%$ to $66.77\%$, indicating that adding ECG to the RESP$+$EEG pair likely lowers recall.
Regarding GAME, the best configuration is ECG$+$EDA$+$RESP, with an average of $66.46\%$—a combination that excludes EEG. Its metrics ($65.91\%$ accuracy, $68.16\%$ precision, $65.32\%$ F1) are balanced across all three dimensions, in contrast to many EEG-containing triples where precision is notably higher than F1. This suggests that the three peripheral signals produce a more consistent representation for the GAME workload.
For MATH, RESP$+$SpO$_2+$EEG achieves the best result, with an average of $66.98\%$ ($66.67\%$ accuracy, $68.89\%$ precision, $65.38\%$ F1), an improvement of $+1.13$ over bimodal, with gains across all three metrics. For ALL, EDA$+$RESP$+$EEG reaches $64.17\%$ ($+0.17$ over bimodal), with accuracy, precision, and F1 closely aligned at $64.02\%$, $64.98\%$, and $63.52\%$ respectively.

\begin{table}
\scriptsize
\caption{Trimodal performance across all task settings.}
\label{table:trimodal}
\begin{center}
\begin{threeparttable}
\begin{tabular}{P{0.5cm} P{2.3cm} P{0.9cm} P{0.9cm} P{0.4cm} P{1.0cm}}
\toprule
\multirow[c]{3}{*}{Task} &
\multirow[c]{3}{*}{Modality} &
\multicolumn{4}{c}{Performance} \\
\cmidrule(lr){3-6}
&  & Accuracy & Precision & F1 &\textit{Average} \\
\midrule
\midrule
IQ &ECG, EDA, RESP                 &64.77 &71.09 &61.97 &\textit{65.94}\\\hdashline
IQ &ECG, EDA, SpO\textsubscript{2} &67.80 &70.29 &66.80 &\textit{68.30}\\\hdashline
IQ &ECG, EDA, EEG                  &67.05 &69.03 &65.76 &\textit{67.28}\\\hdashline
IQ &ECG, RESP, SpO\textsubscript{2}&63.26 &68.61 &60.92 &\textit{64.26}\\\hdashline
IQ &ECG, RESP, EEG                 &68.18 &72.35 &66.77 &\textit{\textbf{69.10}}\\\hdashline
IQ &ECG, SpO\textsubscript{2}, EEG &67.05 &71.27 &65.05 &\textit{67.79}\\\hdashline
IQ &EDA, RESP, SpO\textsubscript{2}&66.67 &70.74 &64.64 &\textit{67.35}\\\hdashline
IQ &EDA, RESP, EEG                 &67.80 &69.97 &64.66 &\textit{67.48}\\\hdashline
IQ &EDA, SpO\textsubscript{2}, EEG &67.80 &72.23 &65.73 &\textit{\underline{68.59}}\\\hdashline
IQ &RESP, SpO\textsubscript{2}, EEG&68.18 &69.91 &67.09 &\textit{68.39}\\\midrule

MATH &ECG, EDA, RESP                 &64.39 &70.75 &61.14 &\textit{65.43}\\\hdashline
MATH &ECG, EDA, SpO\textsubscript{2} &60.98 &62.53 &57.29 &\textit{60.27}\\\hdashline
MATH &ECG, EDA, EEG                  &65.53 &69.64 &63.35 &\textit{\underline{66.17}}\\\hdashline
MATH &ECG, RESP, SpO\textsubscript{2}&64.02 &66.52 &60.07 &\textit{63.54}\\\hdashline
MATH &ECG, RESP, EEG                 &64.02 &69.36 &61.31 &\textit{64.90}\\\hdashline
MATH &ECG, SpO\textsubscript{2}, EEG &64.02 &70.60 &60.71 &\textit{65.11}\\\hdashline
MATH &EDA, RESP, SpO\textsubscript{2}&65.15 &66.16 &62.38 &\textit{64.56}\\\hdashline
MATH &EDA, RESP, EEG                 &64.77 &68.61 &62.71 &\textit{65.36}\\\hdashline
MATH &EDA, SpO\textsubscript{2}, EEG &62.12 &67.79 &58.78 &\textit{62.90}\\\hdashline
MATH &RESP, SpO\textsubscript{2}, EEG&66.67 &68.89 &65.38 &\textit{\textbf{66.98}}\\\midrule

GAME &ECG, EDA, RESP                 &65.91 &68.16 &65.32 &\textit{\textbf{66.46}}\\\hdashline
GAME &ECG, EDA, SpO\textsubscript{2} &62.50 &69.29 &58.89 &\textit{63.56}\\\hdashline
GAME &ECG, EDA, EEG                  &65.53 &68.36 &64.04 &\textit{65.98}\\\hdashline
GAME &ECG, RESP, SpO\textsubscript{2}&64.39 &70.82 &61.24 &\textit{65.48}\\\hdashline
GAME &ECG, RESP, EEG                 &65.15 &70.13 &62.72 &\textit{66.00}\\\hdashline
GAME &ECG, SpO\textsubscript{2}, EEG &64.77 &67.39 &63.52 &\textit{65.23}\\\hdashline
GAME &EDA, RESP, SpO\textsubscript{2}&62.88 &70.39 &59.44 &\textit{64.24}\\\hdashline
GAME &EDA, RESP, EEG                 &65.91 &68.84 &64.27 &\textit{\underline{66.34}}\\\hdashline
GAME &EDA, SpO\textsubscript{2}, EEG &65.91 &67.89 &65.22 &\textit{\underline{66.34}}\\\hdashline
GAME &RESP, SpO\textsubscript{2}, EEG&64.39 &68.09 &62.30 &\textit{64.93}\\\midrule

ALL &ECG, EDA, RESP                 &60.10 &63.04 &57.76 &\textit{60.30}\\\hdashline
ALL &ECG, EDA, SpO\textsubscript{2} &55.56 &58.89 &51.50 &\textit{55.32}\\\hdashline
ALL &ECG, EDA, EEG                  &61.99 &65.23 &59.55 &\textit{62.26}\\\hdashline
ALL &ECG, RESP, SpO\textsubscript{2}&55.68 &59.52 &51.69 &\textit{55.63}\\\hdashline
ALL &ECG, RESP, EEG                 &62.63 &64.52 &61.29 &\textit{62.81}\\\hdashline
ALL &ECG, SpO\textsubscript{2}, EEG &63.64 &64.83 &62.77 &\textit{\underline{63.75}}\\\hdashline
ALL &EDA, RESP, SpO\textsubscript{2}&58.96 &57.72 &56.50 &\textit{57.73}\\\hdashline
ALL &EDA, RESP, EEG                 &64.02 &64.98 &63.52 &\textit{\textbf{64.17}}\\\hdashline
ALL &EDA, SpO\textsubscript{2}, EEG &62.88 &67.93 &60.01 &\textit{63.61}\\\hdashline
ALL &RESP, SpO\textsubscript{2}, EEG&61.99 &63.01 &61.38 &\textit{62.13}\\
\bottomrule
\end{tabular}
\begin{tablenotes}[para,flushleft]
\scriptsize
\end{tablenotes}
\end{threeparttable}
\end{center}
\end{table}

\subsection{Quadmodal Combinations}
Table~\ref{table:quadmodal} reports performance for the $4$-modality combinations. 
For MATH, the best result is obtained by ECG$+$EDA$+$RESP$+$SpO$_2$, with an average of $64.89\%$, the only combination without EEG to rank first at any level across the entire evaluation. Its metrics ($64.02\%$ accuracy, $69.54\%$ precision, $61.10\%$ F1) follow the characteristic MATH profile, with relatively high precision and lower F1. All four EEG-containing quadmodal configurations score below this, ranging from $60.60\%$ to $63.24\%$. At the unimodal, trimodal, and quadmodal levels, EEG does not improve upon physiological-only combinations for MATH, and its inclusion appears to degrade performance for this task.
For IQ, the best quadmodal result is ECG$+$EDA$+$RESP$+$EEG at an average of $68.03\%$, which is $3.97$ points below the bimodal peak of $72.00\%$. Precision falls from $73.49\%$ to $70.75\%$ and F1 from $70.93\%$ to $65.92\%$, indicating that adding ECG and EDA to the RESP$+$EEG pair disrupts the precision-recall balance for abstract reasoning.

\begin{table}
\scriptsize
\caption{Quadmodal performance across all task settings.}
\label{table:quadmodal}
\begin{center}
\begin{threeparttable}
\begin{tabular}{P{0.5cm} P{2.7cm} P{0.8cm} P{0.8cm} P{0.3cm} P{0.9cm}}
\toprule
\multirow[c]{3}{*}{Task} &
\multirow[c]{3}{*}{Modality} &
\multicolumn{4}{c}{Performance} \\
\cmidrule(lr){3-6}
&  & Accuracy & Precision & F1 &\textit{Average} \\
\midrule
\midrule
IQ &ECG, EDA, RESP, SpO\textsubscript{2}      &66.29 &72.33 &64.07 &\textit{\underline{67.56}}\\\hdashline
IQ &ECG, EDA, RESP, EEG                       &67.42 &70.75 &65.92 &\textit{\textbf{68.03}}\\\hdashline
IQ &ECG, EDA, SpO\textsubscript{2}, EEG       &66.67 &68.57 &65.69 &\textit{66.98}\\\hdashline
IQ &ECG, RESP, SpO\textsubscript{2}, EEG      &66.29 &70.58 &64.30 &\textit{67.06}\\\hdashline
IQ &EDA, RESP, SpO\textsubscript{2}, EEG      &65.53 &69.16 &63.30 &\textit{66.00}\\\midrule

MATH &ECG, EDA, RESP, SpO\textsubscript{2}    &64.02 &69.54 &61.10 &\textit{\textbf{64.89}}\\\hdashline
MATH &ECG, EDA, RESP, EEG                     &60.23 &67.67 &55.91 &\textit{61.27}\\\hdashline
MATH &ECG, EDA, SpO\textsubscript{2}, EEG     &59.09 &69.05 &53.65 &\textit{60.60}\\\hdashline
MATH &ECG, RESP, SpO\textsubscript{2}, EEG    &61.74 &70.10 &57.87 &\textit{\underline{63.24}}\\\hdashline
MATH &EDA, RESP, SpO\textsubscript{2}, EEG    &59.85 &69.23 &54.74 &\textit{61.27}\\\midrule

GAME &ECG, EDA, RESP, SpO\textsubscript{2}    &62.88 &62.08 &58.29 &\textit{61.08}\\\hdashline
GAME &ECG, EDA, RESP, EEG                     &66.29 &65.32 &64.14 &\textit{65.25}\\\hdashline
GAME &ECG, EDA, SpO\textsubscript{2}, EEG     &67.42 &71.45 &65.91 &\textit{\underline{68.26}}\\\hdashline
GAME &ECG, RESP, SpO\textsubscript{2}, EEG    &68.18 &71.83 &66.41 &\textit{\textbf{68.81}}\\\hdashline
GAME &EDA, RESP, SpO\textsubscript{2}, EEG    &66.29 &68.83 &64.57 &\textit{66.56}\\\midrule

ALL &ECG, EDA, RESP, SpO\textsubscript{2}     &57.83 &60.19 &55.30 &\textit{57.77}\\\hdashline
ALL &ECG, EDA, RESP, EEG                      &57.32 &61.09 &53.73 &\textit{57.38}\\\hdashline
ALL &ECG, EDA, SpO\textsubscript{2}, EEG      &60.35 &59.61 &58.01 &\textit{\underline{59.32}}\\\hdashline
ALL &ECG, RESP, SpO\textsubscript{2}, EEG     &61.24 &61.95 &60.70 &\textit{\textbf{61.30}}\\\hdashline
ALL &EDA, RESP, SpO\textsubscript{2}, EEG     &57.83 &60.99 &54.91 &\textit{57.91}\\
\bottomrule
\end{tabular}
\begin{tablenotes}[para,flushleft]
\scriptsize
\end{tablenotes}
\end{threeparttable}
\end{center}
\end{table}

\subsection{Pentamodal Combination}

Table~\ref{table:pentamodal} reports results for all $5$-modality combinations. The full pentamodal configuration achieves the highest \textit{Average} across the entire evaluation: $73.02\%$ for IQ, with all three individual metrics elevated ($72.35\%$ accuracy, $76.27\%$ precision, $70.43\%$ F1). 
This is the only configuration across all $31$ combinations in which precision exceeds $76\%$, suggesting that the full signal set enables more balanced classification across both workload levels. The gain over the best bimodal result ($72.00\%$) is $+1.02$ points, driven primarily by a jump in precision from $73.49\%$ to $76.27\%$. For GAME, the pentamodal result is $71.15\%$ ($70.83\%$ accuracy, $72.62\%$ precision, $69.99\%$ F1), below the unimodal EEG result of $72.27\%$, confirming that no combination surpasses EEG alone for this task. For MATH ($65.15\%$) and ALL ($63.01\%$), differences from the quadmodal bests are small ($+0.26$ and $+1.71$ points), consistent with the limited sensitivity of these tasks to the number of modalities included. The pentamodal configuration that yields the best cross-task results, despite intermediate combinations often performing worse than unimodal EEG, suggests that the benefit of the full signal set is a property of the complete combination rather than a cumulative effect of adding modalities one by one. Partial combinations may introduce less effective representations in the shared latent space, where dominant channels are diluted without sufficient complementary information from the others.

\begin{table}
\scriptsize
\caption{Pentamodal performance across all task settings.}
\label{table:pentamodal}
\begin{center}
\begin{threeparttable}
\begin{tabular}{P{0.5cm} P{3.1cm} P{0.7cm} P{0.7cm} P{0.3cm} P{0.8cm}}
\toprule
\multirow[c]{3}{*}{Task} &
\multirow[c]{3}{*}{Modality} &
\multicolumn{4}{c}{Performance} \\
\cmidrule(lr){3-6}
&  & Accuracy & Precision & F1 &\textit{Average} \\
\midrule
\midrule
IQ   &ECG, EDA, RESP, SpO\textsubscript{2}, EEG &72.35 &76.27 &70.43 &\textit{\textbf{73.02}}\\\hdashline
MATH &ECG, EDA, RESP, SpO\textsubscript{2}, EEG &64.39 &69.00 &62.06 &\textit{65.15}\\\hdashline
GAME &ECG, EDA, RESP, SpO\textsubscript{2}, EEG &70.83 &72.62 &69.99 &\textit{71.15}\\\hdashline
ALL  &ECG, EDA, RESP, SpO\textsubscript{2}, EEG &62.88 &64.18 &61.98 &\textit{63.01}\\
\bottomrule
\end{tabular}
\begin{tablenotes}[para,flushleft]
\scriptsize
\end{tablenotes}
\end{threeparttable}
\end{center}
\end{table}

\begin{figure*}
\begin{center}
\includegraphics[scale=0.60]{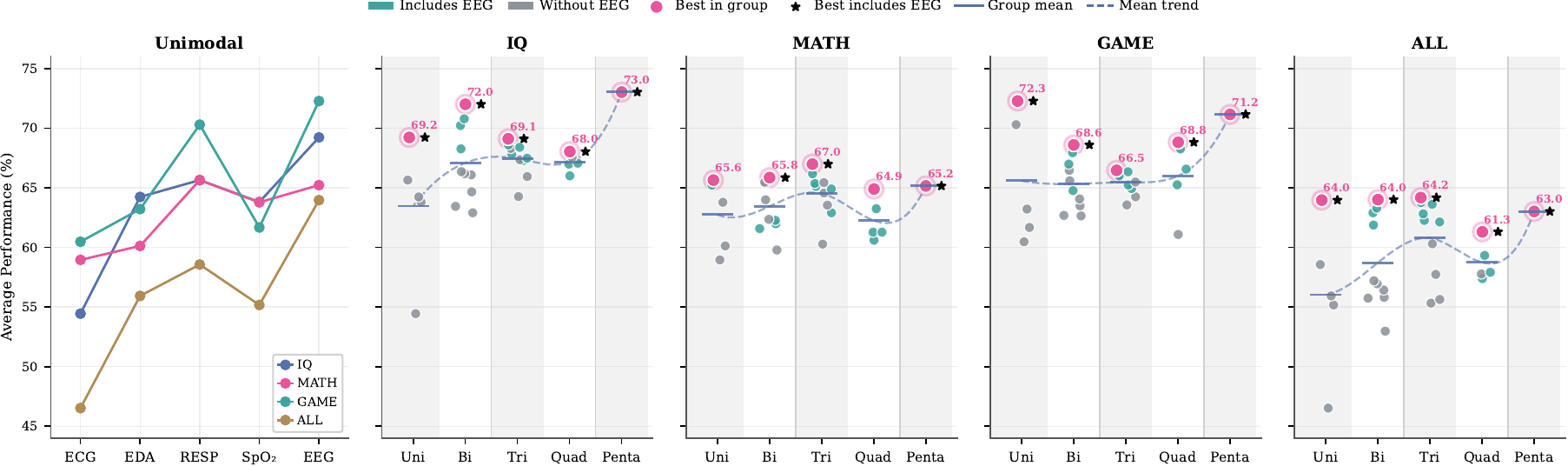}
\end{center}
\caption{Per-task \textit{Average} performance (arithmetic mean of Accuracy, Precision, and F1) across all $31$ modality combinations. \textbf{Left panel}: unimodal \textit{Average} per modality for each task setting (IQ, MATH, GAME, ALL). \textbf{Panels 2--5}: strip plots of all combinations from unimodal through pentamodal, one panel per task. Each dot represents one combination: teal dots include EEG, grey dots do not. Pink dots mark the best-performing combination per level; $\bigstar$ indicates that the best combination includes EEG. Horizontal bars denote the group means; the dashed line shows the mean trend across levels.}
\label{fig:performances_full}
\end{figure*}

\subsection{Overall Analysis \& Discussion}
Figure \ref{fig:performances_full} summarizes the main trends across all $31$ modality combinations. EEG is the strongest single modality, achieving the highest \textit{Average} in IQ, GAME, and ALL, while RESP performs slightly better for MATH. The ALL setting consistently produces lower scores, indicating that pooling different task types increases within-class variability and makes cross-task workload recognition more difficult.
The effect of multimodal fusion is task-dependent. For IQ, the full pentamodal configuration achieves the highest score across the entire evaluation ($73.02\%$). For GAME, however, unimodal EEG remains the best configuration, suggesting that additional peripheral signals may dilute rather than improve the task-relevant EEG representation. For MATH, EEG and non-EEG combinations perform similarly, indicating limited added value from brain activity for this task. These findings show that adding modalities does not guarantee improved performance and that the optimal sensor set depends on the cognitive task.
When averaged across all four settings, the full pentamodal configuration yields the highest cross-task \textit{Average} ($68.08\%$), closely followed by unimodal EEG ($67.67\%$). Thus, the pentamodal model provides the best overall descriptive performance, but the gain over EEG alone is small. Wilcoxon tests further suggest that the advantage of the full set of modalities should be interpreted with caution, given the small cohort size. Overall, the results indicate that EEG is the most reliable single modality, while the proposed unified framework can exploit all modalities when complementary information is available. The channel-stack fusion design also reduces model size and inference time compared with late-fusion alternatives, supporting its suitability for resource-constrained workload assessment.


\section{Conclusion}
This pilot study presented a unified, modality-agnostic hierarchical transformer framework for cognitive workload assessment and evaluated it across all possible combinations of five physiological modalities: ECG, EDA, RESP, SpO\textsubscript{2}, and EEG, using a leave-one-subject-out protocol. Overall, the results show that EEG is the strongest single modality in most settings. Adding more modalities does not consistently improve performance, as intermediate combinations often introduce interference. The full five-modality configuration achieves the highest cross-task average ($68.08\%$) and the highest single-task average on the abstract reasoning task ($73.02\%$), demonstrating that the proposed unified framework can leverage complementary physiological signals without task-specific design. Beyond performance, the channel stack fusion approach requires approximately half as many parameters as embedding-level late-fusion alternatives and consistently achieves lower inference latency across all modalities, making it a practical choice for resource-constrained deployments. The small cohort ($n{=}11$) and single dataset scope reflect the pilot nature of the study; broader validation across larger and more diverse populations, targeted ablations of the latent bottleneck, Fourier encoding, and cross-attention design, and subjective workload modeling using NASA-TLX remain open directions for future work.



\section*{Ethical Impact Statement}
The experimental procedures were approved by the Institutional Review Board. Written informed consent was obtained from all participants prior to the experiment, and no personal information was recorded. The study involved $11$ participants, including $6$ males and $5$ females, aged $20$ to $40$ years, with a mean age of $25 \pm 5.5$ years. Participants did not report neurological conditions and were not under the influence of substances known to impair nervous system function, such as alcohol or nicotine, as disclosed on the day of the experiment.
This study presents a framework for cognitive workload assessment from physiological signals, with potential applications in adaptive human--machine systems and safety-critical environments. The cohort size, controlled laboratory setting, consumer-grade EEG device, and single-dataset protocol limit generalizability to broader populations and real-world settings. Binary easy/hard labels reflect controlled task-difficulty conditions rather than direct subjective workload labels. NASA-TLX scores were collected as subjective workload verification, but their distributions and correlations with model outputs were not analyzed in this study.
The reported results should therefore be interpreted as preliminary. Individual physiological variability, sensor placement, recording quality, interpolation-based temporal alignment, and environmental factors may affect performance in practice. Deployment in applied settings requires validation in larger, more diverse populations, across varied experimental protocols, and through prospective studies assessing reliability across demographic groups. Automated workload assessment also carries risks in high-stakes decision-making without human oversight, since misclassification could lead to inappropriate interventions or failure to respond to elevated cognitive demand. Any future deployment should involve domain experts, ethicists, and end users in the design process.

\section*{Acknowledgments}
The authors used large language model (LLM)-based tools for language editing and improvement. All scientific content, results, and conclusions are solely the work of the authors.

\bibliographystyle{IEEEtran}
\bibliography{library}

\appendix
\section{Appendix}

\subsection{Details \& Complementary Experiments}

\begin{figure}
\begin{center}
\includegraphics[scale=0.55]{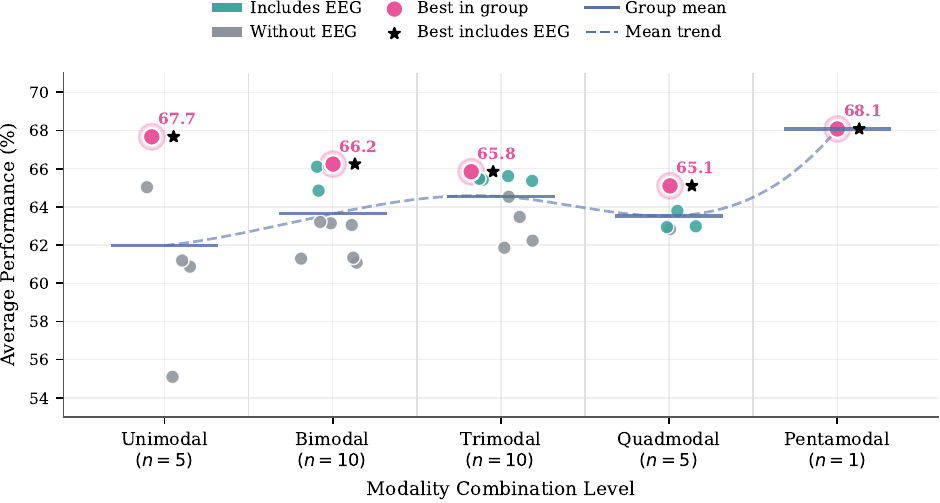}
\end{center}
\caption{Cross-task \textit{Average} performance for all $31$ modality combinations, with each entry averaged over IQ, MATH, GAME, and ALL. Full numerical results are in Table~\ref{table:comparison_modalities} (Appendix).}
\label{fig:performances_avg}
\end{figure}

\begin{figure*}[t]
\begin{center}
\includegraphics[scale=0.70]{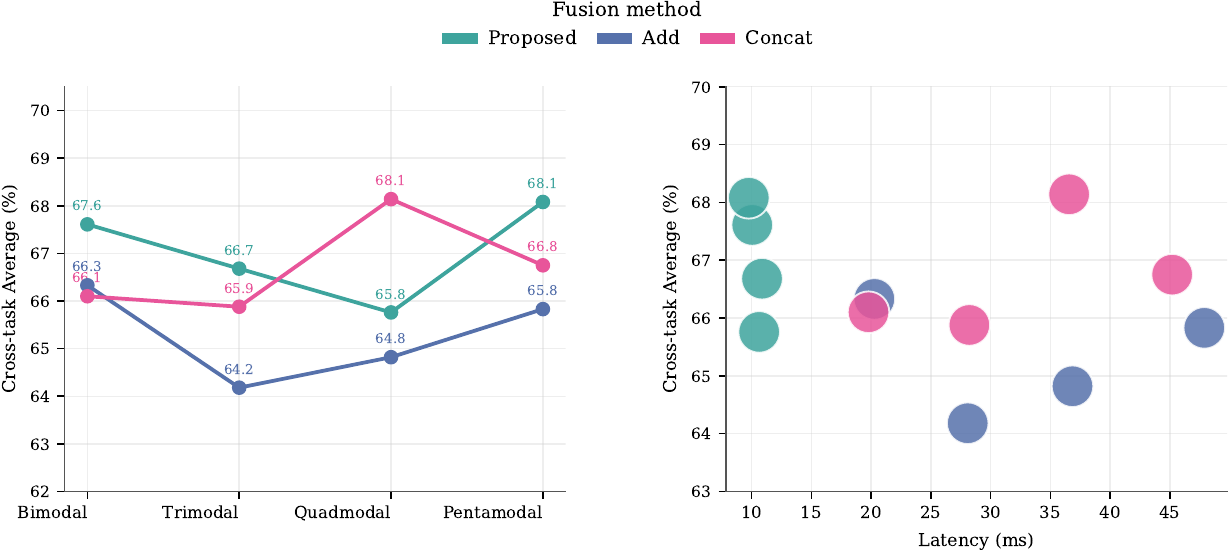}
\end{center}
\caption{Comparison of the proposed channel-stack fusion against late-fusion alternatives (add, concat). \textbf{Left}: cross-task \textit{Average} per modality level. \textbf{Right}: cross-task \textit{Average} versus inference latency. Values correspond to Table \ref{table:comparison_fusion_avg}.}
\label{fig:fusion_cost}
\end{figure*}

Figure~\ref{fig:performances_avg} reports the cross-task \textit{Average} performance for all $31$ modality combinations, with each entry averaged over IQ, MATH, GAME, and ALL. Tables~\ref{table:architecture} and~\ref{table:training} report the per-block architecture hyperparameters and the full training configuration used across all experiments. Table~\ref{table:comparison_modalities} provides the corresponding numerical results for Figure~\ref{fig:performances_avg}, extending the main results by reporting the cross-task \textit{Average} for every modality combination.
Figure~\ref{fig:fusion_cost} and Table~\ref{table:comparison_fusion_avg} compare the proposed channel-stack fusion against two standard late-fusion strategies, namely element-wise addition and concatenation of modality embeddings, across bimodal through pentamodal configurations. Results in Table~\ref{table:comparison_fusion_avg} are reported for the best-performing modality combination at each level. A key structural difference is that late-fusion methods require two separate model instances, one for the multichannel EEG and one for the single-channel peripheral signals, whereas the proposed method processes all modalities through a single shared architecture by stacking them along the channel dimension. This design difference directly affects computational and inference costs: the proposed method requires approximately half as many parameters ($5.6$M vs.\ $11.2$M) and consistently achieves lower latency across all modality levels, as visualized in Figure~\ref{fig:fusion_cost}.
Table~\ref{table:wilcoxon} reports the results of the Wilcoxon signed-rank test comparing EEG and RESP across the four task settings. Table~\ref{table:baseline_comparison} provides an additional comparison with previous EEG-only methods evaluated on the same pilot dataset and LOSO protocol, including \textit{EEGNet} and \textit{1BT}.

\begin{table}[h]
\caption{Per-block architecture hyperparameters.}
\label{table:architecture}
\begin{center}
\begin{tabular}{P{4.2cm} P{3.0cm}}
\toprule
Parameter & Value \\
\midrule
\midrule
Block 1: latent vectors $M_1$, dim $d_1$         & 32,\ 128          \\\hdashline
Block 2: latent vectors $M_2$, dim $d_2$         & 28,\ 112          \\\hdashline
Block 3: latent vectors $M_3$, dim $d_3$         & 24,\ 96           \\\hdashline
Block 4: latent vectors $M_4$, dim $d_4$         & 16,\ 80           \\\hdashline
Cross-attention heads $H_{\text{cr}}$             & 1                 \\\hdashline
Cross-attention head dim $d_{\text{cr}}$          & 64,\ 48,\ 32,\ 16 \\\hdashline
Self-attention heads $H_{\text{la}}$              & 8,\ 6,\ 4,\ 2     \\\hdashline
Self-attention head dim $d_{\text{la}}$           & 64,\ 48,\ 32,\ 16 \\\hdashline
Self-attention rounds per block $R$               & 8,\ 6,\ 4,\ 2     \\\hdashline
Fourier frequency bands $K$                       & 6                 \\\hdashline
Maximum frequency $f_{\max}$                      & 10                \\
\bottomrule
\end{tabular}
\end{center}
\end{table}

\begin{table}[h]
\caption{Augmentation, regularization, and training configuration.}
\label{table:training}
\begin{center}
\begin{tabular}{P{2.8cm} P{4.5cm}}
\toprule
Parameter & Value \\
\midrule
\midrule
\textit{Polarity inversion}  & $p \sim \mathcal{U}(0,\,0.05)$ \\\hdashline
\textit{Additive noise}      & $p \sim \mathcal{U}(0,\,0.05)$, $\alpha \sim \mathcal{U}(1,\,1000)$ \\\hdashline
\textit{Temporal masking}    & $p \sim \mathcal{U}(0,\,0.05)$, $f \sim \mathcal{U}(0.15,\,0.30)$ \\\hdashline
\textit{Label smoothing} $\varepsilon_{\text{ls}}$ & 0.10 \\\hdashline
\textit{Attention dropout}   & 0.10 \\\hdashline
\textit{FF dropout}          & 0.10 \\
\midrule
\midrule
Optimizer                    & AdamW \\\hdashline
Learning rate $\eta$         & $1\times10^{-4}$ \\\hdashline
LR schedule                  & Cosine annealing \\\hdashline
Weight decay $\lambda$       & $0.05$ \\\hdashline
Warm-up epochs               & 20 \\\hdashline
Main epochs                  & 170 \\\hdashline
Cooldown epochs              & 10 \\\hdashline
Total epochs                 & 200 \\\hdashline
Batch size                   & 32 \\
\bottomrule
\end{tabular}
\end{center}
\end{table}

\begin{table}
\scriptsize
\caption{Average Performance per modality or combination of modalities across the 4 tasks.}
\label{table:comparison_modalities}
\begin{center}
\begin{threeparttable}
\begin{tabular}{P{3.1cm} P{0.9cm} P{0.9cm} P{0.9cm} P{0.9cm}}
\toprule
\multirow[c]{3}{*}{Modality} &
\multicolumn{4}{c}{Performance} \\
\cmidrule(lr){2-5}
& Accuracy & Precision & F1 &\textit{Average} \\
\midrule
\midrule
ECG                                        &58.15{\fontsize{5}{6}\selectfont\,\textpm 5.20} &55.53{\fontsize{5}{6}\selectfont\,\textpm 14.20} &51.61{\fontsize{5}{6}\selectfont\,\textpm 10.16} &\textit{55.10}\\\hdashline
EDA                                        &60.89{\fontsize{5}{6}\selectfont\,\textpm 5.60} &64.98{\fontsize{5}{6}\selectfont\,\textpm 10.41} &56.75{\fontsize{5}{6}\selectfont\,\textpm 8.82} &\textit{60.87}\\\hdashline
RESP                                       &64.52{\fontsize{5}{6}\selectfont\,\textpm 6.43} &69.06{\fontsize{5}{6}\selectfont\,\textpm 8.04} &61.52{\fontsize{5}{6}\selectfont\,\textpm 9.51} &\textit{65.03}\\\hdashline
SpO\textsubscript{2}                       &61.24{\fontsize{5}{6}\selectfont\,\textpm 6.42} &64.48{\fontsize{5}{6}\selectfont\,\textpm 10.56} &57.84{\fontsize{5}{6}\selectfont\,\textpm 9.03} &\textit{61.19}\\\hdashline
EEG                                        &67.17{\fontsize{5}{6}\selectfont\,\textpm 6.66} &70.09{\fontsize{5}{6}\selectfont\,\textpm 6.42} &65.75{\fontsize{5}{6}\selectfont\,\textpm 8.02} &\textit{\underline{67.67}}\\\midrule

ECG, EDA                                   &61.56{\fontsize{5}{6}\selectfont\,\textpm 6.16} &64.41{\fontsize{5}{6}\selectfont\,\textpm 10.80} &58.06{\fontsize{5}{6}\selectfont\,\textpm 9.39} &\textit{61.34}\\\hdashline
ECG, RESP                                  &62.34{\fontsize{5}{6}\selectfont\,\textpm 6.08} &67.32{\fontsize{5}{6}\selectfont\,\textpm 6.41} &59.48{\fontsize{5}{6}\selectfont\,\textpm 8.45} &\textit{63.05}\\\hdashline
ECG, SpO\textsubscript{2}                  &61.21{\fontsize{5}{6}\selectfont\,\textpm 5.44} &65.33{\fontsize{5}{6}\selectfont\,\textpm 8.92} &57.32{\fontsize{5}{6}\selectfont\,\textpm 8.68} &\textit{61.29}\\\hdashline
ECG, EEG                                   &66.29{\fontsize{5}{6}\selectfont\,\textpm 8.54} &68.37{\fontsize{5}{6}\selectfont\,\textpm 11.63} &64.05{\fontsize{5}{6}\selectfont\,\textpm 11.14} &\textit{66.24}\\\hdashline
EDA, RESP                                  &62.35{\fontsize{5}{6}\selectfont\,\textpm 5.81} &67.77{\fontsize{5}{6}\selectfont\,\textpm 7.29} &59.31{\fontsize{5}{6}\selectfont\,\textpm 8.31} &\textit{63.14}\\\hdashline
EDA, SpO\textsubscript{2}                  &60.89{\fontsize{5}{6}\selectfont\,\textpm 4.36} &64.92{\fontsize{5}{6}\selectfont\,\textpm 7.74} &57.44{\fontsize{5}{6}\selectfont\,\textpm 7.86} &\textit{61.08}\\\hdashline
EDA, EEG                                   &65.50{\fontsize{5}{6}\selectfont\,\textpm 8.79} &66.87{\fontsize{5}{6}\selectfont\,\textpm 14.23} &62.18{\fontsize{5}{6}\selectfont\,\textpm 12.28} &\textit{64.85}\\\hdashline
RESP, SpO\textsubscript{2}                 &62.79{\fontsize{5}{6}\selectfont\,\textpm 6.28} &66.36{\fontsize{5}{6}\selectfont\,\textpm 7.00} &60.47{\fontsize{5}{6}\selectfont\,\textpm 8.33} &\textit{63.21}\\\hdashline
RESP, EEG                                  &65.94{\fontsize{5}{6}\selectfont\,\textpm 7.77} &68.68{\fontsize{5}{6}\selectfont\,\textpm 9.68} &63.78{\fontsize{5}{6}\selectfont\,\textpm 9.44} &\textit{66.13}\\\hdashline
SpO\textsubscript{2}, EEG                  &65.66{\fontsize{5}{6}\selectfont\,\textpm 8.15} &69.67{\fontsize{5}{6}\selectfont\,\textpm 9.83} &62.96{\fontsize{5}{6}\selectfont\,\textpm 10.39} &\textit{66.10}\\\midrule

ECG, EDA, RESP                             &63.79{\fontsize{5}{6}\selectfont\,\textpm 5.42} &68.26{\fontsize{5}{6}\selectfont\,\textpm 6.71} &61.55{\fontsize{5}{6}\selectfont\,\textpm 7.11} &\textit{64.53}\\\hdashline
ECG, EDA, SpO\textsubscript{2}             &61.71{\fontsize{5}{6}\selectfont\,\textpm 5.31} &65.25{\fontsize{5}{6}\selectfont\,\textpm 7.99} &58.62{\fontsize{5}{6}\selectfont\,\textpm 7.84} &\textit{61.86}\\\hdashline
ECG, EDA, EEG                              &65.02{\fontsize{5}{6}\selectfont\,\textpm 5.93} &68.07{\fontsize{5}{6}\selectfont\,\textpm 5.61} &63.18{\fontsize{5}{6}\selectfont\,\textpm 8.58} &\textit{65.42}\\\hdashline
ECG, RESP, SpO\textsubscript{2}            &61.84{\fontsize{5}{6}\selectfont\,\textpm 6.91} &66.37{\fontsize{5}{6}\selectfont\,\textpm 9.78} &58.48{\fontsize{5}{6}\selectfont\,\textpm 10.10} &\textit{62.23}\\\hdashline
ECG, RESP, EEG                             &65.00{\fontsize{5}{6}\selectfont\,\textpm 5.88} &69.09{\fontsize{5}{6}\selectfont\,\textpm 6.98} &63.02{\fontsize{5}{6}\selectfont\,\textpm 7.29} &\textit{65.70}\\\hdashline
ECG, SpO\textsubscript{2}, EEG             &64.87{\fontsize{5}{6}\selectfont\,\textpm 6.47} &68.52{\fontsize{5}{6}\selectfont\,\textpm 7.18} &63.01{\fontsize{5}{6}\selectfont\,\textpm 7.82} &\textit{65.47}\\\hdashline
EDA, RESP, SpO\textsubscript{2}            &63.42{\fontsize{5}{6}\selectfont\,\textpm 6.19} &66.25{\fontsize{5}{6}\selectfont\,\textpm 10.00} &60.74{\fontsize{5}{6}\selectfont\,\textpm 9.12} &\textit{63.47}\\\hdashline
EDA, RESP, EEG                             &65.62{\fontsize{5}{6}\selectfont\,\textpm 6.18} &68.10{\fontsize{5}{6}\selectfont\,\textpm 8.07} &63.79{\fontsize{5}{6}\selectfont\,\textpm 8.52} &\textit{65.84}\\\hdashline
EDA, SpO\textsubscript{2}, EEG             &64.68{\fontsize{5}{6}\selectfont\,\textpm 5.87} &68.96{\fontsize{5}{6}\selectfont\,\textpm 6.74} &62.44{\fontsize{5}{6}\selectfont\,\textpm 7.69} &\textit{65.36}\\\hdashline
RESP, SpO\textsubscript{2}, EEG            &65.31{\fontsize{5}{6}\selectfont\,\textpm 6.41} &67.48{\fontsize{5}{6}\selectfont\,\textpm 6.53} &64.04{\fontsize{5}{6}\selectfont\,\textpm 7.41} &\textit{65.61}\\\midrule

ECG, EDA, RESP, SpO\textsubscript{2}       &62.76{\fontsize{5}{6}\selectfont\,\textpm 5.85} &66.04{\fontsize{5}{6}\selectfont\,\textpm 9.25} &59.69{\fontsize{5}{6}\selectfont\,\textpm 8.56} &\textit{62.83}\\\hdashline
ECG, EDA, RESP, EEG                        &62.82{\fontsize{5}{6}\selectfont\,\textpm 5.31} &66.21{\fontsize{5}{6}\selectfont\,\textpm 8.71} &59.93{\fontsize{5}{6}\selectfont\,\textpm 8.18} &\textit{62.98}\\\hdashline
ECG, EDA, SpO\textsubscript{2}, EEG        &63.36{\fontsize{5}{6}\selectfont\,\textpm 4.30} &67.17{\fontsize{5}{6}\selectfont\,\textpm 7.42} &60.82{\fontsize{5}{6}\selectfont\,\textpm 6.73} &\textit{63.79}\\\hdashline
ECG, RESP, SpO\textsubscript{2}, EEG       &64.36{\fontsize{5}{6}\selectfont\,\textpm 6.16} &68.62{\fontsize{5}{6}\selectfont\,\textpm 6.98} &62.32{\fontsize{5}{6}\selectfont\,\textpm 7.78} &\textit{65.10}\\\hdashline
EDA, RESP, SpO\textsubscript{2}, EEG       &62.38{\fontsize{5}{6}\selectfont\,\textpm 6.62} &67.05{\fontsize{5}{6}\selectfont\,\textpm 7.69} &59.38{\fontsize{5}{6}\selectfont\,\textpm 9.21} &\textit{62.94}\\\midrule
ECG, EDA, RESP, SpO\textsubscript{2}, EEG  &67.61{\fontsize{5}{6}\selectfont\,\textpm 7.52} &70.52{\fontsize{5}{6}\selectfont\,\textpm 7.00} &66.12{\fontsize{5}{6}\selectfont\,\textpm 9.26} &\textit{\textbf{68.08}}\\
\bottomrule
\end{tabular}
\begin{tablenotes}[para,flushleft]
\scriptsize
\end{tablenotes}
\end{threeparttable}
\end{center}
\end{table}

\begin{table*}[h]
\scriptsize
\caption{Cross-task \textit{Average} performance, computational cost, and inference cost of late-fusion methods (add, concat) versus the proposed channel-stack fusion, across modality levels. For each level and task, the best-performing modality combination is selected, and its \textit{Average} (arithmetic mean of Accuracy, Precision, and F1) is reported. Cross-task \textit{Average} is the arithmetic mean across the four task settings (IQ, MATH, GAME, ALL). Computational and inference costs are measured on an NVIDIA RTX 4090 GPU with batch size 1.}
\label{table:comparison_fusion_avg}
\begin{center}
\begin{threeparttable}
\begin{tabular}{P{1.1cm} P{1.1cm} P{0.7cm} P{0.7cm} P{0.7cm} P{0.7cm} P{1.2cm} P{1.4cm} P{1.1cm} P{2.4cm} P{2.0cm}}
\toprule
\multirow[c]{3}{*}{\makecell{Modality \\ Level}} &
\multirow[c]{3}{*}{Fusion} &
\multicolumn{4}{c}{Task \textit{Average}} &
\multirow[c]{3}{*}{\makecell{Cross-task \\ \textit{Average}}} &
\multicolumn{2}{c}{Computational Cost} &
\multicolumn{2}{c}{Inference Cost} \\
\cmidrule(lr){3-6} \cmidrule(lr){8-9}\cmidrule(lr){10-11}
& & IQ & MATH & GAME & ALL &  & Params (M) & GFLOPs &Latency (ms) GPU$\downarrow$  &Samples/s GPU$\uparrow$ \\
\midrule
\midrule
\multirow{3}{*}{Bimodal}
  & proposed & 72.00 & 65.85 & 68.59 & 64.00 & 67.61 &5.60 &0.35  &10.05 &99.52\\\cdashline{2-11}
  & add      & 69.60 & 66.54 & 65.27 & 63.91 & 66.33 &11.20 &0.69 &20.27 &49.33\\\cdashline{2-11}
  & concat   & 68.24 & 64.25 & 69.15 & 62.76 & 66.10 &11.20 &0.69 &19.78 &50.55\\
\midrule
\multirow{3}{*}{Trimodal}
  & proposed & 69.10 & 66.98 & 66.46 & 64.17 & 66.68 &5.60 &0.35  &10.87 &92.00 \\\cdashline{2-11}
  & add      & 68.10 & 62.52 & 61.61 & 64.47 & 64.18 &11.20 &1.03 &28.08 &35.61 \\\cdashline{2-11}
  & concat   & 65.47 & 69.37 & 62.84 & 65.82 & 65.88 &11.20 &1.03 &28.22 &35.43 \\
\midrule
\multirow{3}{*}{Quadmodal}
  & proposed & 68.03 & 64.89 & 68.81 & 61.30 & 65.76 &5.60 &0.35  &10.63  &64.08 \\\cdashline{2-11}
  & add      & 66.97 & 61.89 & 66.38 & 64.02 & 64.82 &11.20 &1.37 &36.85 &27.14 \\\cdashline{2-11}
  & concat   & 71.66 & 68.54 & 65.42 & 66.92 & 68.14 &11.20 &1.37 &36.57 &27.35 \\
\midrule
\multirow{3}{*}{Pentamodal}
  & proposed & 73.02 & 65.15 & 71.15 & 63.01 & 68.08 &5.61 &0.36  &9.76 &102.48 \\\cdashline{2-11}
  & add      & 68.15 & 62.76 & 65.84 & 66.57 & 65.83 &11.20 &1.71 &47.88 &20.89 \\\cdashline{2-11}
  & concat   & 66.86 & 66.10 & 67.77 & 66.26 & 66.75 &11.20 &1.71 &45.19 &22.13 \\
\bottomrule
\end{tabular}
\begin{tablenotes}[para,flushleft]
\scriptsize
Per-task values are the \textit{Average} (arithmetic mean of Accuracy, Precision, F1) for the best-performing modality combination at each level, taken from the main results tables (proposed) and the fusion table (add/concat). For the late-fusion methods, Params and GFLOPs reflect the combined cost of both model instances. Latency is the mean per-batch inference time; Samples/s is the throughput. $\downarrow$ lower is better; $\uparrow$ higher is better.
\end{tablenotes}
\end{threeparttable}
\end{center}
\end{table*}

\begin{table}
\scriptsize
\caption{Wilcoxon signed-rank test: EEG vs.\ RESP.}
\label{table:wilcoxon}
\begin{center}
\begin{threeparttable}
\begin{tabular}{P{0.7cm} P{0.9cm} P{0.9cm} P{0.4cm} P{0.6cm} P{0.4cm} P{1.4cm}}
\toprule
Task & EEG & RESP & $W$ & $p$ & $r$ & Significance \\
\midrule
\midrule
IQ   &68.18 &64.39 &9.0  &0.109 &0.80 &---  \\\hdashline
MATH &64.77 &64.77 &21.5 &0.905 &0.52 &---  \\\hdashline
GAME &71.97 &69.32 &25.5 &0.838 &0.54 &---  \\\hdashline
ALL  &63.76 &59.60 &6.0 &0.014 &0.91 &\textbf{\checkmark} \\
\bottomrule
\end{tabular}
\begin{tablenotes}[para,flushleft]
\scriptsize
Accuracy (\%) values are taken from Table~\ref{table:single_half}. $r$: rank-biserial correlation (effect size).
\end{tablenotes}
\end{threeparttable}
\end{center}
\end{table}

\begin{table}
\scriptsize
\caption{Comparison with previous EEG-only methods evaluated on the same pilot dataset using the LOSO protocol.}
\label{table:baseline_comparison}
\begin{center}
\begin{threeparttable}
\begin{tabular}{P{1.4cm} P{2.0cm} P{1.0cm} P{1.0cm} P{1.2cm}}
\toprule
Method & Modality & Params & GFLOPs & Mean Avg.\\
\midrule
EEGNet\cite{lawhern_solon_2018} & EEG & 0.01M & 0.02 & 64.47 \\\hdashline
1BT\cite{gkikas_cruz_eeite_cwl_2026} & EEG & 0.45M & 0.02 & 69.37 \\\hdashline
Proposed & ECG, EDA, RESP, SpO\textsubscript{2}, EEG & 5.61M & 0.36 & \textbf{69.77} \\
\bottomrule
\end{tabular}
\begin{tablenotes}[para,flushleft]
\scriptsize
\textit{Mean Avg.} is computed only over IQ, MATH, and GAME, since previous EEG-only methods do not report the pooled ALL setting.
\end{tablenotes}
\end{threeparttable}
\end{center}
\end{table}


\end{document}